\documentclass[a4paper,11pt]{article}

\usepackage{authblk}
\usepackage{fancyhdr}
\usepackage{algorithm}
\usepackage{algorithmic}
\usepackage{multirow}
\usepackage{graphicx}
\usepackage{booktabs}
\usepackage{epsfig}
\usepackage{epstopdf}
\usepackage{amsmath}
\usepackage{caption}
\usepackage{subfig}
\usepackage{amsfonts}
\usepackage{wasysym}
\usepackage{paralist}
\usepackage{url}
\usepackage{float}
\usepackage{booktabs}
\usepackage{todonotes}
\usepackage{microtype}
\usepackage{tikz}
\usepackage{tikz-qtree}

\DeclareMathOperator{\fit}{fit}

\providecommand{\keywords}[1]{\textbf{\textit{Keywords }} #1}

\begin{document}

\title{CoInGP: Convolutional Inpainting with Genetic Programming}

\author[1]{Domagoj Jakobovic}
\author[2]{Luca Manzoni}
\author[3]{Luca Mariot}
\author[3]{Stjepan Picek}
\author[4]{Mauro Castelli}

\affil[1]{{\normalsize Faculty of Electrical Engineering and Computing,
    University of Zagreb, Unska 3, Zagreb, Croatia} \\
	
	{\small \texttt{domagoj.jakobovic@fer.hr}}}
	
\affil[2]{{\normalsize Dipartimento di Matematica e Geoscienze,
Università degli Studi di Trieste, Via Valerio 12/1, Trieste, Italy} \\
	
	{\small \texttt{lmanzoni@units.it}}}

\affil[3]{{\normalsize Cyber Security Research Group, Delft University of Technology, Mekelweg 2, Delft, The Netherlands} \\

  {\small \texttt{\{l.mariot, s.picek\}@tudelft.nl}}}
  
\affil[4]{{\normalsize NOVA Information Management School (NOVA IMS), Universidade Nova de Lisboa, Campus de Campolide, 1070-312, Lisbon, Portugal} \\
	
	{\small \texttt{mcastelli@novaims.unl.pt}}}

\maketitle

\begin{abstract}
  We investigate the use of Genetic Programming (GP) as a convolutional predictor for missing pixels in images. The training phase is performed by sweeping a  sliding window over an image, where the pixels on the border represent the inputs of a GP tree. The output of the tree is taken as the predicted value for the central pixel. We consider two topologies for the sliding window, namely the \emph{Moore} and the \emph{Von Neumann neighborhood}. The best GP tree scoring the lowest prediction error over the training set is then used to predict the pixels in the test set. We experimentally assess our approach through two experiments. In the first one, we train a GP tree over a subset of $1000$ complete images from the MNIST dataset. The results show that GP can learn the distribution of the pixels with respect to a simple baseline predictor, with no significant differences observed between the two neighborhoods. In the second experiment, we train a GP convolutional predictor on two degraded images, removing around $20\%$ of their pixels. In this case, we observe that the Moore neighborhood works better, although the Von Neumann neighborhood allows for a larger training set.
\end{abstract}


\thispagestyle{fancy}

\section{Introduction}
\label{sec:intro}

Nowadays, images represent a common testbed to evaluate the performance of many algorithms, especially those coming from the deep learning domain~\cite{ILSVRC15,isola2016imagetoimage,janke2019analysis,chen2018looks}. The usability of images in this context is impaired if they are damaged or incomplete. Indeed, missing pixels can severely impact the information carried by the images and hinder the performances of artificial intelligence techniques trained on them. Hence, there is often the need to resort to image inpainting techniques. \emph{Digital inpainting} generally denotes all methods related to the reconstruction of lost or damaged parts of an image by means of algorithms that replace such parts. We refer the reader to the recent surveys by Elharrous et al.~\cite{elharrouss20} and Jam et al.~\cite{jam21} for a more complete overview of image inpainting techniques, while in the following, we recall only the essential approaches investigated in this research field.

Traditionally, two techniques have been explored for the image inpainting procedure. \emph{Exemplar-based methods} fill a missing region by exploiting local information in the surrounding area. This can be done both at the level of single pixels, as in the pioneering work by Efros and Leung~\cite{efros1999}, or patch-wise, by searching for replacement patches in the parts of the image that are not damaged, as proposed for instance by Criminisi et al.~\cite{criminisi04}. On the other hand, in \emph{diffusion-based techniques} inpainting is performed by spreading the image information from the boundary of a missing region towards its center, an approach that was initially investigated by Bertalmio et al.~\cite{bertalmio00}. A further research thread also focused on combining both the exemplar-based and diffusion-based approaches by defining \emph{hybrid methods}, as done for instance in~\cite{bertalmio03}.

More recently, \emph{deep learning methods}, and in particular convolutional neural networks (CNNs), have shown excellent results on image inpainting tasks due to their ability to use large training sets~\cite{liu2018image}. The part where CNNs truly have an advantage over other inpainting techniques is the fact that they can better capture the global structure of an image~\cite{10.1007/978-3-030-01264-9_1}. Finally, researchers also used generative adversarial networks (GANs) for many image-to-image translation tasks, including image inpainting~\cite{isola2016imagetoimage}.

When considering evolutionary algorithms, there are not many works examining the image inpainting task. Li et al. used a combination of a total variation method and a genetic algorithm for completing an image~\cite{Li2016ImageIA}. Li and Yang proposed a patch-based method based on evolutionary algorithms that search for the optimal patch in the area around the damaged region~\cite{6631705}. Interestingly, while convolutional neural networks represent state-of-the-art in image translation tasks, up to now, there are not many attempts to employ the convolutional paradigm in other artificial intelligence techniques. To the best of our knowledge, there is only a single work that considers how to combine convolutions and genetic programming~\cite{10.1007/978-3-030-21077-9_5}. There, the authors applied their method to develop image denoising filters with a multi-layer architecture.

This paper proposes a novel technique for the image inpainting task based on Genetic Programming (GP)~\cite{10.5555/138936} and convolutions. We denote our approach as CoInGP -- Convolutional Inpainting with Genetic Programming. Our technique works locally by considering the immediate neighbors of a missing pixel, which are used as the input of a GP tree. The output evaluated at the root of the tree represents the predicted value for the central missing pixel. The window is then slid over the image, and the prediction process is repeated for the remaining missing pixels, thus obtaining a reconstructed image. We tackle the problem of evolving a suitable GP tree as a supervised learning task over known pixels. In particular, the training set is composed of fitness cases where the inputs are the values of the neighboring pixels for a specific position of the window, while the label corresponds to the correct value of the central pixel. The optimization objective consists in minimizing the RMSE between the predictions made by the GP tree and the correct labels over all fitness cases.

As far as we are aware, this is the first paper considering GP for image inpainting. Hence, more than comparing with state-of-the-art deep learning methods such as CNNs and GANs (which we leave for future research), the main motivation of our work is to search for preliminary evidence that convolutional inpainting can also be performed with Genetic Programming as an underlying learning primitive. Incidentally, we adopted a similar approach in~\cite{gecco2020} for the domain of automatic text generation. For these reasons, we frame the investigation presented in this paper around two general research questions:
\begin{enumerate}
    \item Can CoInGP learn the distribution of the pixels' intensities in a dataset of \emph{complete} images?
    \item Can CoInGP obtain a plausible reconstruction of a \emph{single} degraded image by training on the available pixels?
\end{enumerate}

For the first research question, we perform the training on a subset of $1000$ images from the MNIST dataset~\cite{deng2012mnist} without missing pixels. The fitness of a GP tree in the population is evaluated by predicting the value of each pixel in all selected images (excluding those at borders, which do not have enough neighbors). The best evolved GP tree is then independently validated on another test set of $1000$ complete images from MNIST. Concerning the second research question, we conduct an experiment on two different test images, where we remove around $20\%$ of the pixels. In this case, the training is done on the available pixels, while the testing phase consists in predicting the actual missing pixels.

Further, we investigate a third research question that is orthogonal to the previous two: namely, whether the shape of the sliding window plays a role in the performance of GP when predicting the central pixel. To this end, we consider two different topologies for the window: \emph{Moore neighborhood} and \emph{Von Neumann neighborhood}.

Since this paper is mostly an empirical investigation of our approach's feasibility, in all our experiments, we compare the results obtained by CoInGP against those achieved by a simple baseline method, i.e., the predictor that computes the average value of the pixels in the neighborhood.% In particular, we leave the comparison with other state-of-the-art approaches in the inpainting literature for future research.

Our findings can be summarized as follows: regarding the first research question, GP is indeed able to learn the distribution of the pixels in a dataset of complete images to a certain extent, since for both neighborhood shapes, the evolved trees obtain a significantly lower RMSE than the respective baseline predictor. Moreover, in this case, we observe no statistically significant difference between Moore and Von Neumann neighborhoods. We obtain similar results for the second research question since CoInGP reaches a lower RMSE value than the baseline predictor when reconstructing the missing pixels of the two test images. However, in this case, there is a further difference between the two topologies considered for the sliding window, with Moore neighborhood achieving a better performance. This finding is especially interesting since, for geometrical reasons, Moore neighborhood can exploit a smaller training set than the Von Neumann neighborhood.

The rest of this paper is organized as follows. Section~\ref{sec:problem} formalizes the problem of predicting the central pixel in a sliding window by exploiting the information in the surrounding ones. Section~\ref{sec:gp-conv} presents the details of our CoInGP method, showing how a GP tree can be used to predict an image's pixels and defining an appropriate fitness function to evaluate the quality of its predictions. Section~\ref{sec:exp-set} describes the experimental settings adopted in our empirical assessment of CoInGP and summarizes the obtained results. Section~\ref{sec:discussion} gives an interpretation of the main experimental findings that can be drawn from our results and formulates some hypotheses worth exploring to investigate the observed behavior of CoInGP further. Finally, Section~\ref{sec:conclusions} recaps the main contributions of our paper and suggests future research directions on the subject.

\section{Problem Formulation}
\label{sec:problem}
This section formalizes the problem of predicting pixels in an image, which will be tackled with genetic programming in the remainder of the paper. In what follows, we consider an input image as a matrix $I$ of size $M\times N$, where each entry $x_{(i,j)}$ is the intensity value of the pixel at coordinates $(i,j)$ for $i \in [M]$ and $j \in [N]$, where $[M] = \{1,\cdots,M\}$ and $[N] = \{1,\cdots,N\}$. For illustration purposes, we deal only with $8$-bit greyscale images, so that each entry $x_{(i,j)}$ in the matrix is an integer number between $0$ and $255$; nevertheless, our approach can be generalized to any color depth.

Suppose that the image is \emph{damaged}, that is, the intensities of a subset of $k$ of its pixels $S = \{(i_1,j_1), \cdots (i_k, j_k)\} \subseteq [M]\times[N]$ are missing. The goal is to recover the original intensities $x_{(i_1,j_1)}, \cdots x_{(i_k, j_k)}$ starting from those that are still available, i.e., the pixels in the complementary set $P = [M]\times [N] \setminus S$. This task is also known as \emph{inpainting} in the image processing literature~\cite{bugeau2010,guillemot2014}. One of the possible approaches to perform inpainting stands on the fundamental observation that the intensities of neighboring pixels are \emph{correlated}. In a probabilistic framework, this property can also be restated as the fact that the probability distribution of a pixel's intensity given the intensities of the pixels in its neighborhood is independent of the rest of the image~\cite{efros1999}.

This observation suggests that, to recover the intensity of a missing pixel in an image, one can use just the values of its neighboring pixels as an input for the prediction. More formally, the two main topologies that can be adopted are the \emph{Moore} neighborhood and the \emph{Von Neumann} neighborhood~\cite{schiff2011}. Considering only neighborhoods of radius $1$ (i.e., only the immediate neighbors of a pixels are taken into account), for the Moore neighborhood the input to predict a pixel in position $(i,j)$ will be a $3 \times 3$ matrix defined as:
\begin{equation}
    \label{eq:moore}
\mathcal{N}_{i,j}=
\begin{bmatrix}
    x_{(i-1,j-1)} & x_{(i-1,j)} & x_{(i-1,j+1)} \\
    x_{(i,j-1)} &  & x_{(i,j+1)} \\
    x_{(i+1,j-1)} & x_{(i+1,j)} & x_{(i+1,j+1)} \\
  \end{bmatrix} \enspace ,
\end{equation} 
where the $8$ elements on the border represent the intensities of the pixels in the neighborhoods, and the goal is to predict the value of the central pixel. Analogously, for a Von Neumann neighborhood the input to the prediction will be the following matrix:
\begin{equation}
    \label{eq:von}
\mathcal{N}_{i,j}=
\begin{bmatrix}
     & x_{(i-1,j)} &  \\
    x_{(i,j-1)} &  & x_{(i,j+1)} \\
     & x_{(i+1,j)} &  \\
  \end{bmatrix} \enspace ,
\end{equation}
where, in this case, we do not consider the elements in the corners and the input for predicting the central pixel are only the four elements which are respectively at its top, bottom, left, and right.

Intuitively, the quality of the prediction will also depend upon the number of available neighboring pixels: in particular, if also some of the neighboring pixels of $\mathcal{N}_{i,j}$ are missing in the degraded image, then we will have less information at our disposal to predict the central pixel $x_{(i,j)}$. In what follows, we adopt the simplifying assumption that each missing pixel in the degraded image is ``sufficiently far'' from all other missing pixels, or equivalently that each missing pixel has a complete neighborhood. Formally, in the case of Moore neighborhood this means that the \emph{Chebyshev distance} $d_\infty$ between any pair of missing pixels $(i_{t_1}, j_{t_1}),(i_{t_2}, j_{t_2}) \in S$ must be strictly greater than $1$:
\begin{displaymath}
d_\infty((i_{t_1}, j_{t_1}),(i_{t_2}, j_{t_2})) = \max\{|i_{t_1} - i_{t_2}|, |j_{t_1}-j_{t_2}|\} > 1 \enspace .
\end{displaymath}
Analogously, for the Von Neumann neighborhood the constraint is that the \emph{Manhattan distance} $d_1$ between $(i_{t_1}, j_{t_1})$ and $(i_{t_2}, j_{t_2})$ has to be greater than $1$:
\begin{displaymath}
d_1((i_{t_1}, j_{t_1}),(i_{t_2}, j_{t_2})) = |i_{t_1} - i_{t_2}| + |j_{t_1}-j_{t_2}| > 1 \enspace .
\end{displaymath}

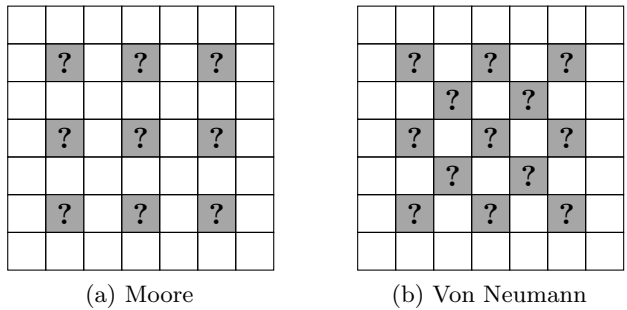
\begin{figure}[h]
  \centering
    \subfloat[Moore\label{fig:moore-ex}]{
    \begin{tikzpicture}
      [->,auto,node distance=1.5cm, empt node/.style={font=\sffamily,inner
        sep=0pt,minimum size=0.3cm}, rect
        node/.style={rectangle,draw,font=\bfseries,minimum size=0.5cm, inner
        sep=0pt, outer sep=0pt}, grey
        node/.style={rectangle,draw,fill=gray!70,font=\bfseries,minimum size=0.5cm, inner sep=0pt, outer sep=0pt}]

    \node [grey node] (m1) {?};
    \node [rect node] (m2) [right=0cm of m1] {};
    \node [grey node] (m3) [right=0cm of m2] {?};
    \node [rect node] (m4) [right=0cm of m3] {};
    \node [rect node] (m5) [left=0cm of m1] {};
    \node [grey node] (m6) [left=0cm of m5] {?};
    \node [rect node] (m7) [left=0cm of m6] {};
    \node [rect node] (a1) [above=0cm of m1] {};
    \node [rect node] (a2) [right=0cm of a1] {};
    \node [rect node] (a3) [right=0cm of a2] {};
    \node [rect node] (a4) [right=0cm of a3] {};
    \node [rect node] (a5) [left=0cm of a1] {};
    \node [rect node] (a6) [left=0cm of a5] {};
    \node [rect node] (a7) [left=0cm of a6] {};
    \node [grey node] (b1) [above=0cm of a1] {?};
    \node [rect node] (b2) [right=0cm of b1] {};
    \node [grey node] (b3) [right=0cm of b2] {?};
    \node [rect node] (b4) [right=0cm of b3] {};
    \node [rect node] (b5) [left=0cm of b1] {};
    \node [grey node] (b6) [left=0cm of b5] {?};
    \node [rect node] (b7) [left=0cm of b6] {};
    \node [rect node] (c1) [above=0cm of b1] {};
    \node [rect node] (c2) [right=0cm of c1] {};
    \node [rect node] (c3) [right=0cm of c2] {};
    \node [rect node] (c4) [right=0cm of c3] {};
    \node [rect node] (c5) [left=0cm of c1] {};
    \node [rect node] (c6) [left=0cm of c5] {};
    \node [rect node] (c7) [left=0cm of c6] {};
    \node [rect node] (d1) [below=0cm of m1] {};
    \node [rect node] (d2) [right=0cm of d1] {};
    \node [rect node] (d3) [right=0cm of d2] {};
    \node [rect node] (d4) [right=0cm of d3] {};
    \node [rect node] (d5) [left=0cm of d1] {};
    \node [rect node] (d6) [left=0cm of d5] {};
    \node [rect node] (d7) [left=0cm of d6] {};
    \node [grey node] (e1) [below=0cm of d1] {?};
    \node [rect node] (e2) [right=0cm of e1] {};
    \node [grey node] (e3) [right=0cm of e2] {?};
    \node [rect node] (e4) [right=0cm of e3] {};
    \node [rect node] (e5) [left=0cm of e1] {};
    \node [grey node] (e6) [left=0cm of e5] {?};
    \node [rect node] (e7) [left=0cm of e6] {};
    \node [rect node] (f1) [below=0cm of e1] {};
    \node [rect node] (f2) [right=0cm of f1] {};
    \node [rect node] (f3) [right=0cm of f2] {};
    \node [rect node] (f4) [right=0cm of f3] {};
    \node [rect node] (f5) [left=0cm of f1] {};
    \node [rect node] (f6) [left=0cm of f5] {};
    \node [rect node] (f7) [left=0cm of f6] {};
    
    \end{tikzpicture}
    }
    \phantom{MM}
    \subfloat[Von Neumann\label{fig:ex-rev-orb}]{
    \begin{tikzpicture}
      [->,auto,node distance=1.5cm, empt node/.style={font=\sffamily,inner
        sep=0pt,minimum size=0.3cm}, rect
        node/.style={rectangle,draw,font=\bfseries,minimum size=0.5cm, inner
        sep=0pt, outer sep=0pt}, grey
        node/.style={rectangle,draw,fill=gray!70,font=\bfseries,minimum size=0.5cm, inner sep=0pt, outer sep=0pt}]

    \node [grey node] (m1) {?};
    \node [rect node] (m2) [right=0cm of m1] {};
    \node [grey node] (m3) [right=0cm of m2] {?};
    \node [rect node] (m4) [right=0cm of m3] {};
    \node [rect node] (m5) [left=0cm of m1] {};
    \node [grey node] (m6) [left=0cm of m5] {?};
    \node [rect node] (m7) [left=0cm of m6] {};
    \node [rect node] (a1) [above=0cm of m1] {};
    \node [grey node] (a2) [right=0cm of a1] {?};
    \node [rect node] (a3) [right=0cm of a2] {};
    \node [rect node] (a4) [right=0cm of a3] {};
    \node [grey node] (a5) [left=0cm of a1] {?};
    \node [rect node] (a6) [left=0cm of a5] {};
    \node [rect node] (a7) [left=0cm of a6] {};
    \node [grey node] (b1) [above=0cm of a1] {?};
    \node [rect node] (b2) [right=0cm of b1] {};
    \node [grey node] (b3) [right=0cm of b2] {?};
    \node [rect node] (b4) [right=0cm of b3] {};
    \node [rect node] (b5) [left=0cm of b1] {};
    \node [grey node] (b6) [left=0cm of b5] {?};
    \node [rect node] (b7) [left=0cm of b6] {};
    \node [rect node] (c1) [above=0cm of b1] {};
    \node [rect node] (c2) [right=0cm of c1] {};
    \node [rect node] (c3) [right=0cm of c2] {};
    \node [rect node] (c4) [right=0cm of c3] {};
    \node [rect node] (c5) [left=0cm of c1] {};
    \node [rect node] (c6) [left=0cm of c5] {};
    \node [rect node] (c7) [left=0cm of c6] {};
    \node [rect node] (d1) [below=0cm of m1] {};
    \node [grey node] (d2) [right=0cm of d1] {?};
    \node [rect node] (d3) [right=0cm of d2] {};
    \node [rect node] (d4) [right=0cm of d3] {};
    \node [grey node] (d5) [left=0cm of d1] {?};
    \node [rect node] (d6) [left=0cm of d5] {};
    \node [rect node] (d7) [left=0cm of d6] {};
    \node [grey node] (e1) [below=0cm of d1] {?};
    \node [rect node] (e2) [right=0cm of e1] {};
    \node [grey node] (e3) [right=0cm of e2] {?};
    \node [rect node] (e4) [right=0cm of e3] {};
    \node [rect node] (e5) [left=0cm of e1] {};
    \node [grey node] (e6) [left=0cm of e5] {?};
    \node [rect node] (e7) [left=0cm of e6] {};
    \node [rect node] (f1) [below=0cm of e1] {};
    \node [rect node] (f2) [right=0cm of f1] {};
    \node [rect node] (f3) [right=0cm of f2] {};
    \node [rect node] (f4) [right=0cm of f3] {};
    \node [rect node] (f5) [left=0cm of f1] {};
    \node [rect node] (f6) [left=0cm of f5] {};
    \node [rect node] (f7) [left=0cm of f6] {};
    
    \end{tikzpicture}
    }
  \caption{Densest packings of missing pixels allowed respectively under unitary Moore and Von Neumann neighborhoods.}
  \label{fig:ex-distance}
\end{figure}

The consequence of these constraints is that missing pixels can share the frontier of the neighborhood under consideration, but a missing pixel cannot be in the \emph{frontier} of another one. In particular, the frontier of a neighborhood of radius $r$ is defined as the set of pixels at a distance $r$ from the central one. Since we are only considering the case of radius $r=1$, the frontier corresponds to the set of all pixels in the neighborhood except the central one. As an example, Figure~\ref{fig:ex-distance} shows the densest packing of missing pixels one can have for the Moore and Von Neumann neighborhood, respectively. The Von Neumann topology allows for more missing pixels under the same image size since it includes fewer neighbors than the Moore topology. Also, observe that for both neighborhoods the missing pixels cannot occur on the border of the image, i.e., $1 < i < M$ and $1 < j < N$ for every missing pixel $(i,j) \in S$.

Although this separation hypothesis does not always hold in realistic scenarios, we decided to adopt it to initially validate the suitability of our method, since as we mentioned before, as far as we are aware, this is the first attempt employing GP to predict missing pixels in images with a convolutional approach.

\section{GP as a Convolutional Predictor}
\label{sec:gp-conv}

The main idea that we investigate in this paper is to evolve GP trees that act as \emph{convolutional operators} to predict the values of missing pixels. Similarly to what is done in \emph{Convolutional Neural Networks} (CNNs)~\cite{Goodfellow-et-al-2016}, we assume that the transformation used to predict the values of the missing pixels is \emph{shift-invariant}. This means there is a \emph{local function} $f$ which is applied over a small \emph{sliding window} of neighboring pixels and is shifted one place at a time over the whole image. The output of the function $f$ corresponds to the predicted intensity of the pixel at the center of the window.

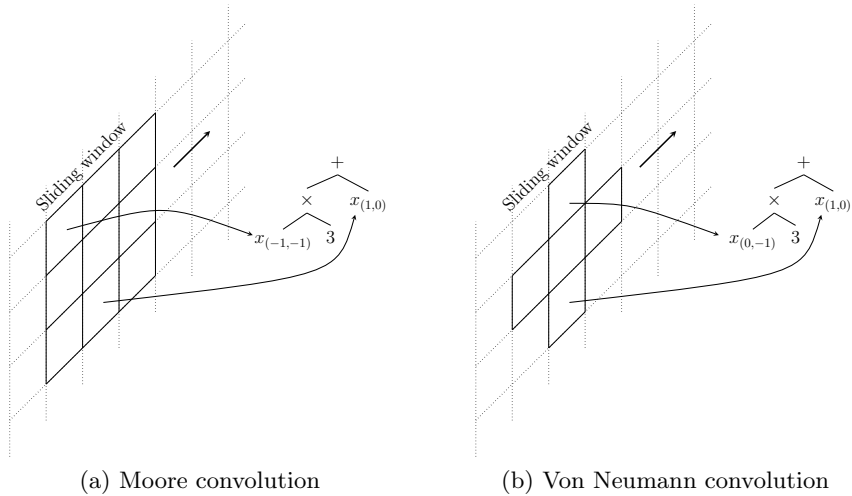
\begin{figure}[h]
  \centering
  \subfloat[Moore convolution\label{fig:tab-cl-moore}]{
  \resizebox{!}{6cm}{
  \begin{tikzpicture}
    % background lines
    \draw[dotted] (-1,-2) -- (-1,4.5);
    \draw[dotted] (0,-1) -- (0,5.5);
    \draw[dotted] (1,0) -- (1,6.5);
    \draw[dotted] (2,1) -- (2,7.5);
    \draw[dotted] (3,2) -- (3,8.5);
    \draw[dotted] (4,3) -- (4,9.5);
    \draw[dotted] (5,4) -- (5,10.5);
    
    \draw[dotted] (-1,-1) -- (5.5,5.5);
    \draw[dotted] (-1,0.5) -- (5.5,7);
    \draw[dotted] (-1,2) -- (5.5,8.5);
    \draw[dotted] (-1,3.5) -- (5.5,10);
    
    % Vertical lines
    \draw[thick] (0,0) -- (0,4.5);
    \draw[thick] (1,1) -- (1,5.5);
    \draw[thick] (2,2) -- (2,6.5);
    \draw[thick] (3,3) -- (3,7.5);
    
    % Horizontal lines
    \draw[thick] (0,0) -- (3,3);
    \draw[thick] (0,1.5) -- (3,4.5);
    \draw[thick] (0,3) -- (3,6);
    \draw[thick] (0,4.5) -- (3,7.5);
    
    % Moving direction
    \draw[very thick,->,shorten >=0pt,shorten <=0pt,>=stealth] (3.5, 6) -- (4.5,7);
    
    % Connecting lines
    \draw[thick,->,shorten >=10pt,shorten <=2pt,>=stealth] (0.5,4.25)  .. controls (3.25,5) ..  (6,4);
    \draw[thick,->,shorten >=2pt,shorten <=2pt,>=stealth] (1.5,2.25) .. controls (8,3) .. (8.5,4.75);
    
    % Text
    \node[rotate=45] [anchor=south west] at (0,4.5) {\Large Sliding window};
    
    % GP Tree
    \begin{scope}[shift={(8,6)}]
      \Large
      \Tree [.$+$ [.$\times$ $x_{(-1,-1)}$ 3 ] $x_{(1,0)}$ ]
    \end{scope}
  \end{tikzpicture}
  }
  }%
  \phantom{MM}%
  \subfloat[Von Neumann convolution\label{fig:tab-cl-vn}]{
  \resizebox{!}{6cm}{
  \begin{tikzpicture}
    % background lines
    \draw[dotted] (-1,-2) -- (-1,4.5);
    \draw[dotted] (0,-1) -- (0,5.5);
    \draw[dotted] (1,0) -- (1,6.5);
    \draw[dotted] (2,1) -- (2,7.5);
    \draw[dotted] (3,2) -- (3,8.5);
    \draw[dotted] (4,3) -- (4,9.5);
    \draw[dotted] (5,4) -- (5,10.5);
    
    \draw[dotted] (-1,-1) -- (5.5,5.5);
    \draw[dotted] (-1,0.5) -- (5.5,7);
    \draw[dotted] (-1,2) -- (5.5,8.5);
    \draw[dotted] (-1,3.5) -- (5.5,10);
    
    % Vertical lines
    \draw[thick] (0,1.5) -- (0,3);
    \draw[thick] (1,1) -- (1,5.5);
    \draw[thick] (2,2) -- (2,6.5);
    \draw[thick] (3,4.5) -- (3,6);
    
    % Horizontal lines
    \draw[thick] (1,1) -- (2,2);
    \draw[thick] (0,1.5) -- (3,4.5);
    \draw[thick] (0,3) -- (3,6);
    \draw[thick] (1,5.5) -- (2,6.5);
    
    % Moving direction
    \draw[very thick,->,shorten >=0pt,shorten <=0pt,>=stealth] (3.5, 6) -- (4.5,7);
    
    % Connecting lines
    \draw[thick,->,shorten >=10pt,shorten <=2pt,>=stealth] (1.5,5)  .. controls (3.25,5) ..  (6,4);
    \draw[thick,->,shorten >=2pt,shorten <=2pt,>=stealth] (1.5,2.25) .. controls (8,3) .. (8.5,4.75);
    
    % Text
    \node[rotate=45] [anchor=south west] at (0,4.5) {\Large Sliding window};
    
    % GP Tree
    \begin{scope}[shift={(8,6)}]
      \Large
      \Tree [.$+$ [.$\times$ $x_{(0,-1)}$ 3 ] $x_{(1,0)}$ ]
    \end{scope}
  \end{tikzpicture}
  }
  }
  \caption{Convolutional prediction based on GP with the Moore and Von Neumann neighborhood of radius $1$. The pixels in the frontier of the neighborhood currently looked by the sliding window are fed as input variables to the GP tree, and its output is taken as the predicted value for the central pixel.}
  \label{fig:convol-gp}
\end{figure}

In our setting, we consider both the case of a square $3\times 3$ sliding window, which corresponds to the Moore neighborhood of radius $1$, and a cross-shaped window of width $3$, which represents the Von Neumann neighborhood of radius $1$. In the former case, the local function has the form $f: [0,255]^8 \to [0,255]$, while in the latter it is $f: [0,255]^4 \to [0,255]$; either way, the local function is expressed with a GP tree. Thus, the $8$ (respectively, $4$) intensities of the pixels on the border of the window are taken as terminal nodes of the GP tree, and the value generated at the root node will be the prediction for the central pixel. Figure~\ref{fig:convol-gp} depicts the idea of using a GP tree as a convolutional predictor by sliding a window over the image for the case of Moore and Von Neumann neighborhoods.

To construct such a convolutional predictor, we need to define an appropriate fitness function that measures how good a particular GP tree is in determining the correct value for the central pixel. The idea is to frame the problem in terms of supervised learning, with the \emph{training set} including fitness cases where the inputs are the values of the pixels in the neighborhood, and the labels are the correct values for the corresponding central pixel. Recall from Section~\ref{sec:intro} that we are interested in two research questions, which translates to the following tasks:

\begin{enumerate}
    \item Given a set of complete images (i.e., without missing pixels) drawn from a common dataset, learn the distribution of the pixels' intensities in this set.
    \item Given a \emph{single} degraded image, reconstruct the complete image by predicting the values of the missing pixels.
\end{enumerate}

For Task (1), let $\mathcal{I} = \{I_1, \cdots, I_n\}$ be a set of images, each of size $M\times N$ and without missing pixels. For each image $I_k$, with $k \in \{1,\cdots,n\}$, we define the corresponding \emph{set of fitness cases} (or training examples) as follows:
\begin{equation}
\label{eq:fitness-cases}
F_k = \{(\mathcal{N}_{i,j}, x_{(i,j)}): \ 1 < i < M, \ 1 < j < M\} \enspace ,
\end{equation}
where $\mathcal{N}_{i,j}$ is the punctured neighborhood matrix defined as in Eqs.~\eqref{eq:moore} and~\eqref{eq:von}, respectively when the Moore and Von Neumann neighborhood is used. In other words, for each pixel $(i,j)$ in image $I_k$ (except for those on the borders), we construct the corresponding neighborhood matrix $\mathcal{N}_{i,j}$ (\emph{without} the value of the pixel in the center) which is used as an input to a GP tree $\tau$. The actual intensity $x_{i,j}$ of the central pixel $(i,j)$ is retained as the correct label of the training example. The total number of fitness cases in $F_k$ is thus $(M-2)(N-2)$. Next, the global training set is defined as the union of the fitness cases sets of all images in $\mathcal{I}$:
\begin{equation}
\label{eq:train-complete}
T_1 = \bigcup_{k=1}^n F_k \enspace .
\end{equation}

For Task (2), we consider a single degraded image $I$ of size $M \times N$, where $S = \{(i_1,j_1), \cdots (i_k, j_k)\}$ is the subset of missing pixels that satisfy respectively the Chebyshev distance $d_\infty > 1$ constraint (if the Moore neighborhood is adopted) or the Manhattan distance $d_1 > 1$ constraint (if the Von Neumann neighborhood is used). Further, let $P = [M] \times [N] \setminus S$ be the complementary subset of available pixels. Then, the training set is defined as follows:
\begin{equation}
\label{eq:training-set}
T_2 = \{(\mathcal{N}_{i,j}, x_{(i,j)}): \ (i,j) \in P, \ 1 < i < M, \ 1 < j < M\} \enspace .
\end{equation}
Hence, $T_2$ is a particular case of Eq.~\eqref{eq:fitness-cases}, where the training examples are constrained only to the available pixels of the image having a complete neighborhood.

Given the output $\hat{x}_{(i,j)} = \tau(\mathcal{N}_{(i,j)})$, we can compute the error that the GP tree $\tau$ made in predicting the correct pixel intensity $x_{(i,j)}$. Generalizing to all available training examples, we define the fitness function for the GP tree $\tau$ as the \emph{root mean square error} (RMSE) over the training set:
\begin{equation}
    \label{eq:fitness}
    \fit(\tau) = \sqrt{\frac{\sum_{(\mathcal{N}_{i,j},x_{(i,j)}) \in T} (\tau(\mathcal{N}_{i,j}) - x_{(i,j)})^2}{|T|}} \enspace .
\end{equation}
Hence, the optimization objective is to minimize $\fit$, since having a GP tree that achieves a small RMSE means that its predictions are close to the actual pixel values. Observe that it is not necessary to specify the precise form of the training set in Eq.~\eqref{eq:fitness} depending on Task (1) or (2), since $T_1$ simply concatenates the training examples of all images in the dataset $\mathcal{I}$.

Once the GP evolution process has terminated, the best individual undergoes a \emph{testing phase}. In Task (1), the best GP tree is used to predict the value of each pixel in all images of a test set $\mathcal{T}$ different from $\mathcal{I}$, although always drawn at random from the same dataset. Conversely, for Task (2), the best tree is used to predict the values of the pixels in the missing set $S$ of the target image $I$. In both cases, the performance of the best tree is evaluated again with the RMSE measure. Clearly, in Task (2), this approach assumes that the missing set $S$ can be artificially created to retain the original values of the pixels in it for computing the RMSE.

\section{Experimental Phase}
\label{sec:exp-set}

This section describes the experimental evaluation that we conducted to investigate the two research questions outlined in Section~\ref{sec:intro} through our CoInGP method. In what follows, we first discuss the common experimental settings and parameters adopted in our study. Then, we describe the setup and the results obtained for our two experiments, namely, learning the distribution of the pixels' intensities for a set of complete images from the MNIST dataset and predicting the missing pixels in two degraded test images\footnote{The source code of our implementation of CoInGP is publicly available at \url{https://github.com/rymoah/CoInGP}}.

\subsection{Common Parameters}
\label{subsec:common}
To experimentally assess our method, we loosely followed the GP parameter settings that we adopted in~\cite{gecco2020} for another supervised learning task, namely next word prediction, and checked with preliminary experiments that they were suitable for the image inpainting task as well. In particular, in each GP run, we evolved a population of $500$ individuals for $500$ generations, which amounts to 250\,000 evaluations. The selection phase was performed using tournament selection with a tournament size of $3$, where the worst individual is replaced by the offspring generated by applying crossover on the best two individuals. For the crossover, we adopted simple subtree, uniform, size fair, one-point, and context preserving crossover, randomly selected at each crossover operation. The newly generated individual undergoes a mutation subject to individual mutation probability of $0.3$; we used a simple subtree mutation~\cite{poli08}. To avoid bloat, we set the maximum tree depth to $8$, which corresponds to the number of input variables available in the Moore neighborhood. The terminal symbols for the GP trees included random constant values in the range $[-1,1]$ and either the $8$ (for Moore neighborhood) or $4$ (for Von Neumann neighborhood) input variables corresponding to the intensities of the available pixels in the respective neighborhood. The functional symbols for the internal nodes are taken from the following set: $\sin$, $\cos$, $+$, $-$, $/$ (protected), $*$, $\min$, $\max$, $\mathrm{avg}$, $\sqrt{\cdot}$ and $\mathrm{pos}$. The square root operator returns zero if the argument is negative, while the unary operator $\mathrm{pos}$ is defined as $\mathrm{pos}(x) = x$ if $x \geq 0$ and $0$ otherwise.

Since we require the predicted pixel intensity to be an integer number between $0$ and $255$, we constrained the output of a GP tree by first clipping it in the interval $[0,255]$ (i.e., if $|\tau(\mathcal{N}_{i,j})| > 255$ we set $|\tau(\mathcal{N}_{i,j})| = 255$), and then by applying a \emph{linear scaling operator} to obtain the closest integer value, using the method proposed by Keijzer~\cite{keijzer2003}. An alternative solution would be to directly use byte-oriented operators in the functional set, such as bitwise logical operations, modular additions, and rotations. However, we deemed that this approach would have constrained too much the search space explored by GP, hindering its ability to generate good tree predictors with low RMSE fitness values.

\subsection{Experiments on the MNIST Dataset}

For the first research question, we considered the well-known MNIST dataset~\cite{deng2012mnist}, which contains images of handwritten digits. In particular, each image has a fixed size of $28\times 28$ pixels, with the digit placed at the center. For each experimental run, we randomly sampled from this dataset $1000$ images for the training set, with the same number of images for each digit, and we constructed the corresponding training set $T_1$ according to Eq.~\eqref{eq:train-complete}, and minimized the RMSE as defined in Eq.~\eqref{eq:fitness}. In total, we performed 30 independent runs. At the end of each run, we validated the best GP tree with another random sample of $1000$ images. The test set is still constructed using Eq.~\eqref{eq:train-complete} and the performance criterion is the minimization of the fitness function. Thus, the idea is to verify whether the GP tree resulting from the training phase can score a small RMSE on a set of unseen images.

\begin{figure}[h]
    \centering
    \includegraphics[scale=0.5]{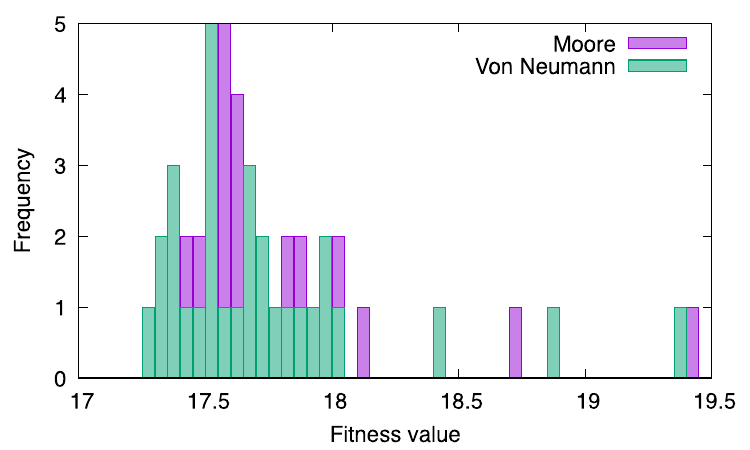}
    \caption{Histograms representing the distribution of the fitness values for the best individuals achieved in the same $30$ independent runs, for both neighborhoods.}
    \label{fig:mnist_result}
\end{figure}

%The results obtained from this experiment are reported in Figure~\ref{fig:mnist_result}.
The obtained results suggest that GP is indeed learning the distribution of the pixels in the training set. Indeed, the convergence of the best fitness during the training phase for the Moore and the Von Neumann neighborhoods showed that the RMSE decreased over all $30$ experimental runs, thus indicating that the predicted pixels are closer and closer to their target values. The plot in Figure~\ref{fig:mnist_result} shows the distribution of the fitness values, on the test set, for both Moore and Von Neumann neighborhood, over the $30$ independent runs. To compare the results obtained on the test phase, we also computed the RMSE for the baseline predictors that replace the central pixel with the average value of the neighboring ones for the images. This resulted in an RMSE of $33.488$ and $27.191$ for the baseline predictors based on the Moore and the Von Neumann neighborhoods, respectively. Based on these results, one can observe that CoInGP is obtaining significantly better results than the baseline method in predicting the pixels over the test set since the RMSE values of the former are in the range $17.25-19.5$ for both neighborhoods. Moreover, the overlapping of the two distributions indicates that the performance of CoInGP is not dependent on the neighborhood's choice. We further validated this qualitative observation through a statistical test. In particular, the Mann-Whitney test was executed (with a significance level of $\alpha=0.05$) under the null hypothesis that the median fitness of the two series of data (i.e., the one using Moore neighborhood and the one using Von Neumann neighborhood) were equal. The obtained $p$-value ($0.6228$) led us to not reject the null hypothesis, thus confirming that there is no difference between the two neighborhoods used by CoInGP.

The obtained results suggest the suitability of the proposed approach for the reconstruction of the damaged pixels of an image. The same results do not highlight a difference between the two neighborhood structures.

\subsection{Experiments on Single Images}

\begin{figure}[h]
    \centering
    \includegraphics[scale=0.5]{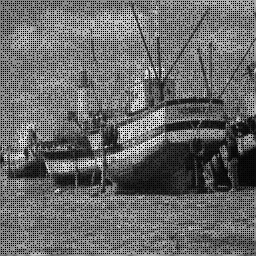}%
    \phantom{M}%
    \includegraphics[scale=0.5]{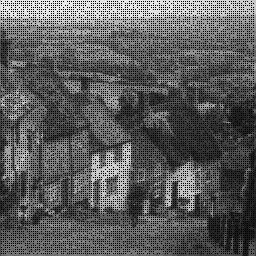}
    \caption{The damaged test images: Boat and Goldhill.}
    \label{fig:damaged-images}
\end{figure}

To validate the previous findings in a more realistic scenario, the second part of the experimental phase applies the proposed approach to images that present a more complex pattern than the MNIST images. We employ two $256 \times 256$ grayscale images on which approximately $20\%$ of the pixels were removed. The two images (with the removed pixels) are presented in Figure~\ref{fig:damaged-images}.

We adopted the following procedure to generate the damaged images: every two columns of the image, the first one was kept unchanged while $100$ non-adjacent pixels were randomly removed from the second. Overall, this procedure resulted in removing $12\,700$ pixels out of $65\,536$, corresponding to a percentage of removed pixels equal to $19.38\%$ for each image. As detailed in Section~\ref{sec:gp-conv}, the training set $T_2$ used in this learning task is composed of all remaining pixels in the degraded image, along with their complete neighborhoods. Due to the different neighborhood shapes considered, the number of fitness cases for the Moore neighborhood was $4,950$, and for the Von Neumann neighborhood was $21,036$. That is, since the Von Neumann neighborhood contains fewer pixels, it also allows to employ a larger number of fitness cases. In this case, the training phase was performed for $100$ independent runs. The testing is then performed by predicting the values of the removed pixels with the best GP individual at the end of each run, i.e., the one achieving the smaller RMSE over the training set.

\begin{figure}[h]
    \centering
    \includegraphics[scale=0.5]{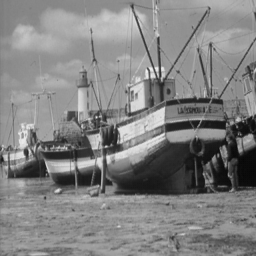}%
    \phantom{M}%
    \includegraphics[scale=0.5]{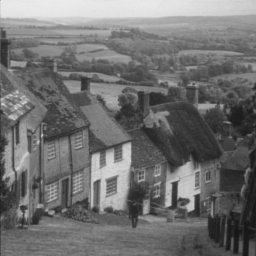}
    \includegraphics[scale=0.5]{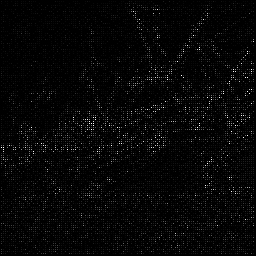}%
    \phantom{M}%
    \includegraphics[scale=0.5]{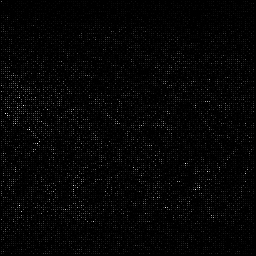}
    \caption{At the top, the images corrected using the Moore neighborhood. At the bottom, the difference, increased ten times, between the reconstructed and the original image.}
    \label{fig:corrected-moore-images}
\end{figure}

\begin{figure}[h]
    \centering
    \includegraphics[scale=0.5]{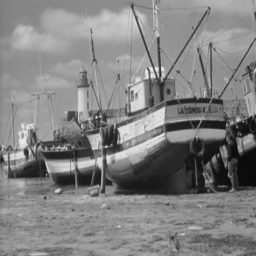}%
    \phantom{M}%
    \includegraphics[scale=0.5]{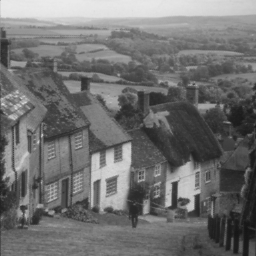}
    \includegraphics[scale=0.5]{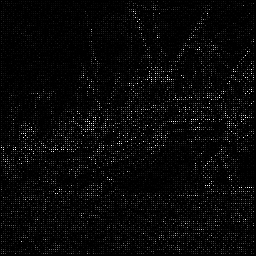}%
    \phantom{M}%
    \includegraphics[scale=0.5]{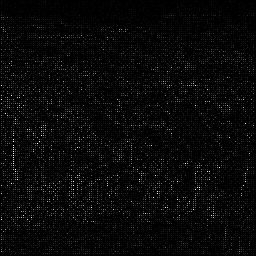}
    \caption{At the top, the images corrected using the Von Neumann neighborhood. At the bottom, the difference, increased ten times, between the reconstructed and the original image.}
    \label{fig:corrected-von-neumann-images}
\end{figure}

\begin{figure}[h]
    \centering
    \includegraphics[scale=0.5]{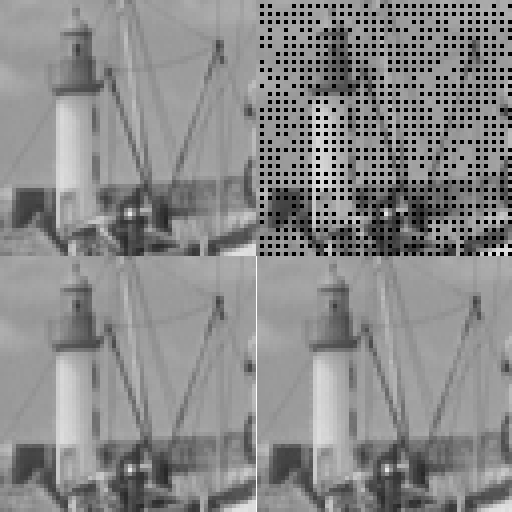}
    \caption{A closeup of the correction performed by GP on one of the images. Clockwise from the top left: original image, damaged image, corrected images with the Von Neumann and Moore neighbourhoods, respectively.}
    \label{fig:comparison-boat}
\end{figure}

The results of the reconstruction process are presented in Figure~\ref{fig:corrected-moore-images} for the Moore neighborhood, and in Figure~\ref{fig:corrected-von-neumann-images} for the Von Neumann neighborhood. A closeup is presented in Figure~\ref{fig:comparison-boat}.

The reconstructed images are both taken from a random GP run. For each image, we also present the pixel-by-pixel difference between the reconstructed image and the original one, where each difference is increased ten times to make it visible. As it is possible to observe, the errors in both cases are limited (i.e., there are no extremely different pixels) and distributed mainly across the edges of the objects in the image. This is particularly visible in the Boat image, where the distribution of the error mostly follows the profiles of the hull and the masts.

\begin{figure}
    \centering
    \includegraphics[scale=0.5]{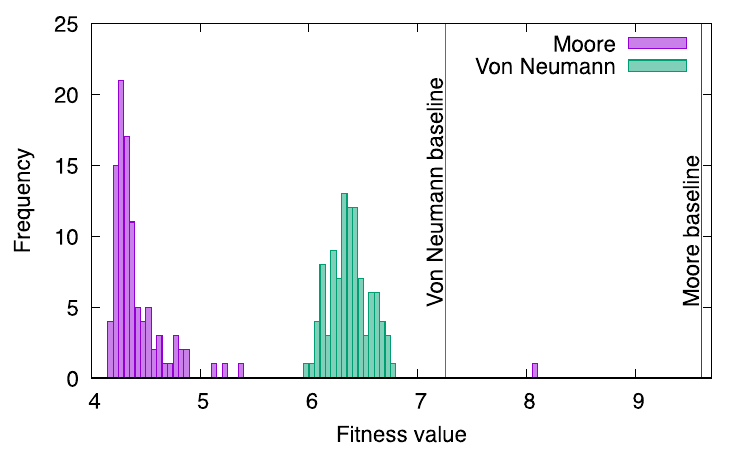}
    \includegraphics[scale=0.5]{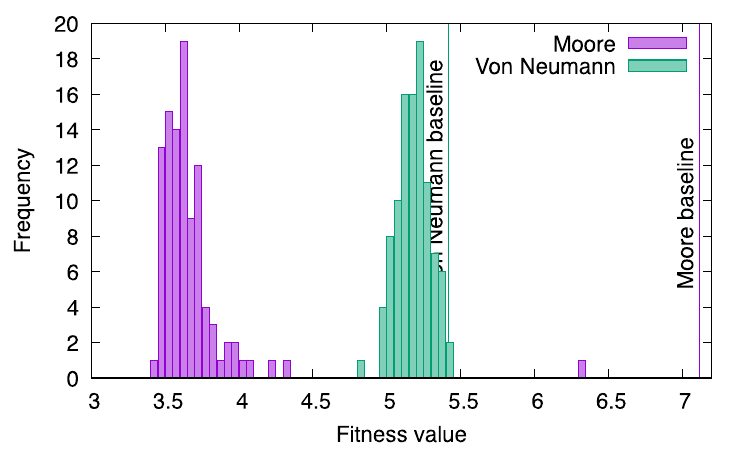}
    \caption{Distribution of best fitness over $100$ runs with both Moore and Von Neumann neighborhoods for the Boat image (top plot) and the Goldhill image (bottom plot).}
    \label{fig:histograms}
\end{figure}

Besides qualitative considerations on the reconstructed images, we also assessed whether CoInGP could predict missing pixels in these images from a quantitative point of view, performing again a comparison with the baseline predictors that compute the average intensities of the neighboring pixels. Figure~\ref{fig:histograms} depicts the plots of the distributions of the best fitness over $100$ experimental runs achieved by GP over each test image.

As a general remark, in most cases, all fitness values obtained are below both baselines, independently of the underlying neighborhood. The only exceptions which occur, however, are limited to a few outliers. In particular, some runs in the distribution of the Moore neighborhood scored an RMSE value between the two baselines, while a small part of the right tail of the Von Neumann distribution overlaps the corresponding baseline in the Goldhill image. In any case, we noticed that the peaks of all GP distributions are significantly distant from the respective baseline fitness values. Further, in all the test images, the use of the Moore neighborhood produces lower fitness values than the Von Neumann neighborhood, even if it allows fewer training samples to be generated.

\section{Discussion}
\label{sec:discussion}

We now interpret the experimental results reported in the previous section in the light of the two research questions stated in Section~\ref{sec:intro}. Regarding the first question, we can empirically conclude that our CoInGP method can successfully learn the distribution of the pixels' intensities in a dataset of complete images, i.e., without missing pixels. Indeed, the convergence plots for the best fitness during the training phase on the MNIST dataset show that the evolutionary process implemented by GP is learning how to minimize the error between the correct label for the central pixel in the window and the predicted one. The distributions of the best fitness on the test set confirm that GP can generalize to unseen images to a certain extent, and a comparison with the baseline predictors shows that it achieves a significantly lower RMSE.

Concerning the second research question, in our experimental setting, the missing pixels accounted for roughly $20\%$ of the pixels of each test image. Our approach's main limitation is that the training process requires a complete neighborhood, i.e., no missing pixels must occur in the frontier of the central pixel whose value has to be predicted. This limits both the number of missing pixels that one can have in the degraded image and their relative positions. However, the preliminary results that we obtained on the test images are promising enough to encourage further improvements in this direction by extending our method to consider the case of \emph{adjacent} missing pixels in the degraded image. An interesting idea to accomplish this task could be to employ a diffusion-based inpainting approach~\cite{7987733}. In this case, the GP predictor would be first convolved on the border of a missing region and then gradually shifted towards its interior.

An interesting difference that can be remarked between the two experiments regards the influence of the sliding window's topology on the performance of CoInGP. In fact, for the MNIST experiment, we detected no significant difference between the Moore and Von Neumann neighborhoods, suggesting that this parameter is not a key factor when learning the distribution of pixels of complete images. Conversely,  when going into the details of the second experiment with a single test image, the GP predictors based on the Moore neighborhood achieved a better performance (i.e., a lower RMSE value) than those using the Von Neumann neighborhood. This happens even though the Von Neumann neighborhood requires fewer input variables to compute the predicted missing pixel and can be optimized on a larger training set. Consequently, this result indicates that GP can learn more efficiently by using a larger number of input variables and a smaller training set. It would be interesting to investigate if this difference in terms of performance also holds for larger neighborhoods. Still, for radius $2$, this would already yield GP trees with $24$ and $12$ input variables, respectively, for the Moore and Von Neumann neighborhood, thereby increasing both the training time and the GP predictors size.

Recall that the baseline predictors simply computed the average of the pixels in the neighborhood to predict the value of the central one. An interesting fact that can be observed from our experiments is that the RMSE achieved by the Von Neumann baseline predictor is lower than that scored by the Moore baseline, both in the MNIST dataset and the single test images. Hence, this suggests that the information for predicting the central pixel is not uniformly distributed across the neighboring ones: it seems that the $4$ ``diagonal'' pixels in the Moore neighborhood contain less information to predict the central one. Nonetheless, this observation is in stark contrast with the fact that GP scored a lower RMSE value with the Moore neighborhood than with the Von Neumann neighborhood. This indicates that CoInGP can learn how to correctly "weigh" the value of the pixels depending on their position. It would be interesting to further investigate this issue by analyzing the structure of the trees evolved by GP with the Moore neighborhood.

Finally, from the qualitative point of view, we observed that the prediction errors made by GP individuals mostly focused around the edges in the test images. This is an expected side effect: if one considers images as two-dimensional spatial signals, edges correspond to \emph{high-frequency} regions, where abrupt changes of the intensity value occur among neighboring pixels. Consequently, the pixels' intensities in a neighborhood where an edge occurs have a lower correlation. Additionally, the independence hypothesis that the probability distribution of a pixel given the surrounding ones is independent of the rest of the image does not hold. This explains why our GP convolutional predictor obtains a higher error on edges' proximity, but it is not necessarily a negative effect: one could use CoInGP to perform edge detection as a by-product. Furthermore, an interesting idea to decrease the prediction error on the edges would be to develop a 2-layer architecture: the first layer would be used to detect the edges, while the second one would perform the inpainting task by discriminating between pixels' types. For the latter case, one could evolve GP trees over a larger neighborhood so that more information can be used to predict the central pixel.

\section{Conclusions and Future Work}
\label{sec:conclusions}
In this paper, we proposed a method for performing convolutional inpainting with GP -- CoInGP. The main idea is to sweep a small sliding window over a degraded image with missing pixels, where the neighborhood pixels captured by the window are fed as input to a GP tree. The GP's output is then taken as the predicted value for the central pixel. The RMSE between the original pixel intensities and those predicted by the GP tree is used to define a fitness function, which has to be minimized. We investigated this approach through two research questions, namely whether GP can learn the distribution of the pixels' intensities from a dataset of complete images and whether GP can restore a plausible reconstructed version of a single degraded image with missing pixels. To this end, we carried out two supervised learning experiments.

In the first experiment, the training set is composed of a random sample of $1000$ images from the MNIST dataset, with the objective of minimizing the RMSE over all pixels of each image. The best GP tree evolved during this phase is then validated by applying it over a distinct test set. The results showed that our CoInGP method was able to generalize to a certain extent on unseen images since it performed better than the respective baseline predictors. Moreover, no difference was observed between using a sliding window with the Moore neighborhood and the Von Neumann neighborhood.

In the second experiment, given a degraded image with missing pixels, an optimal GP tree predictor is evolved by using all available pixels as a training set. For each position of the sliding window, the central pixel is removed and replaced with the value predicted by a GP tree. The test phase consists in applying the best tree evolved by GP on the actual missing pixels. We experimented with two test images. The results showed that GP could evolve trees with better prediction accuracy than the respective baseline predictor. Furthermore, in this case, we observed a clear difference in terms of performance between the Moore and the Von Neumann neighborhood, with the former achieving lower RMSE scores than the latter on the test sets. Considering that the Von Neumann baseline predictor has a lower RMSE than the Moore one, this seems to suggest that GP can learn how to appropriately assess the information contained in the pixels at the corners of the Moore neighborhood.

We conclude by pointing out directions for future research besides those already discussed in the previous section.
The experiments presented in this paper suggest that using GP as a convolutional predictor represents an interesting building block to be plugged in more complex architectures for supervised learning tasks in the image domain. We sketched the first idea of this approach in Section~\ref{sec:discussion}, where we proposed to use a first GP convolutional layer for detecting the edges in an image and then use the second layer to perform inpainting. Thus, it would be interesting to generalize this concept to multiple GP-based convolutional layers and see how the performance of the overall system compares to other analogous and more established methods (i.e., like CNNs). Besides the inpainting technique, one could also consider the application of GP to other image processing tasks that can be formulated as supervised learning problems. This includes not only tasks where the training has to be performed on a single target image, as in the inpainting case, but also on multiple images, such as image classification. In particular, this would likely benefit from the use of a multi-layered architecture where each GP-based convolutional layer would be used to extract a particular feature of an image.

Finally, the convolution strategy is general enough to be applied to any kind of learning task in the signal processing domain. In this paper, we addressed the use case of images, which can be considered as two-dimensional spatial signals, but it could be interesting to explore how convolutional GP behaves on one-dimensional signals such as time series. In particular, the problem of predicting missing data in general signals is also known as \emph{imputation}, which is useful for symbolic regression over incomplete datasets. As far as we know, there are a few works in the literature addressing the imputation problem using GP~\cite{tran2015,tran2017,alhelali2020}, but none of them uses a convolutional approach like the one proposed in this paper. 

\section*{Acknowledgments}
Mauro Castelli acknowledges financial support from Fundação para a Ciência e a Tecnologia (award number: DSAIPA/DS/0022/2018) and from the Slovenian Research Agency (award number: P5-0410).

\bibliographystyle{abbrv}      % mathematics and physical sciences
\bibliography{bibliography}   % name your BibTeX data base

\begin{thebibliography}{10}

\bibitem{alhelali2020}
B.~Al{-}Helali, Q.~Chen, B.~Xue, and M.~Zhang.
\newblock Hessian complexity measure for genetic programming-based imputation
  predictor selection in symbolic regression with incomplete data.
\newblock In {\em Genetic Programming - 23rd European Conference, EuroGP 2020,
  Held as Part of EvoStar 2020, Seville, Spain, April 15-17, 2020,
  Proceedings}, pages 1--17, 2020.

\bibitem{bertalmio00}
M.~Bertalm{\'{\i}}o, G.~Sapiro, V.~Caselles, and C.~Ballester.
\newblock Image inpainting.
\newblock In J.~R. Brown and K.~Akeley, editors, {\em Proceedings of the 27th
  Annual Conference on Computer Graphics and Interactive Techniques, {SIGGRAPH}
  2000, New Orleans, LA, USA, July 23-28, 2000}, pages 417--424. {ACM}, 2000.

\bibitem{bertalmio03}
M.~Bertalm{\'{\i}}o, L.~A. Vese, G.~Sapiro, and S.~J. Osher.
\newblock Simultaneous structure and texture image inpainting.
\newblock {\em {IEEE} Trans. Image Process.}, 12(8):882--889, 2003.

\bibitem{bugeau2010}
A.~Bugeau, M.~Bertalm{\'{\i}}o, V.~Caselles, and G.~Sapiro.
\newblock A comprehensive framework for image inpainting.
\newblock {\em {IEEE} Trans. Image Processing}, 19(10):2634--2645, 2010.

\bibitem{chen2018looks}
C.~Chen, O.~Li, D.~Tao, A.~Barnett, C.~Rudin, and J.~Su.
\newblock This looks like that: Deep learning for interpretable image
  recognition.
\newblock In {\em Advances in Neural Information Processing Systems 32: Annual
  Conference on Neural Information Processing Systems 2019, NeurIPS 2019, 8-14
  December 2019, Vancouver, BC, Canada}, pages 8928--8939, 2019.

\bibitem{criminisi04}
A.~Criminisi, P.~P{\'{e}}rez, and K.~Toyama.
\newblock Region filling and object removal by exemplar-based image inpainting.
\newblock {\em {IEEE} Trans. Image Process.}, 13(9):1200--1212, 2004.

\bibitem{deng2012mnist}
L.~Deng.
\newblock The mnist database of handwritten digit images for machine learning
  research [best of the web].
\newblock {\em IEEE Signal Processing Magazine}, 29(6):141--142, 2012.

\bibitem{efros1999}
A.~A. Efros and T.~K. Leung.
\newblock Texture synthesis by non-parametric sampling.
\newblock In {\em Proceedings of the International Conference on Computer
  Vision, Kerkyra, Corfu, Greece, September 20-25, 1999}, pages 1033--1038,
  1999.

\bibitem{elharrouss20}
O.~Elharrouss, N.~Almaadeed, S.~Al{-}M{\'{a}}adeed, and Y.~Akbari.
\newblock Image inpainting: {A} review.
\newblock {\em Neural Process. Lett.}, 51(2):2007--2028, 2020.

\bibitem{Goodfellow-et-al-2016}
I.~Goodfellow, Y.~Bengio, and A.~Courville.
\newblock {\em Deep Learning}.
\newblock MIT Press, 2016.
\newblock \url{http://www.deeplearningbook.org}.

\bibitem{guillemot2014}
C.~Guillemot and O.~L. Meur.
\newblock Image inpainting : Overview and recent advances.
\newblock {\em {IEEE} Signal Process. Mag.}, 31(1):127--144, 2014.

\bibitem{isola2016imagetoimage}
P.~Isola, J.~Zhu, T.~Zhou, and A.~A. Efros.
\newblock Image-to-image translation with conditional adversarial networks.
\newblock In {\em 2017 {IEEE} Conference on Computer Vision and Pattern
  Recognition, {CVPR} 2017, Honolulu, HI, USA, July 21-26, 2017}, pages
  5967--5976, 2017.

\bibitem{jam21}
J.~Jam, C.~Kendrick, K.~Walker, V.~Drouard, G.~J. Hsu, and M.~H. Yap.
\newblock A comprehensive review of past and present image inpainting methods.
\newblock {\em Comput. Vis. Image Underst.}, 203:103147, 2021.

\bibitem{janke2019analysis}
J.~Janke, M.~Castelli, and A.~Popovi{\v{c}}.
\newblock Analysis of the proficiency of fully connected neural networks in the
  process of classifying digital images. benchmark of different classification
  algorithms on high-level image features from convolutional layers.
\newblock {\em Expert Systems with Applications}, 135:12--38, 2019.

\bibitem{keijzer2003}
M.~Keijzer.
\newblock Improving symbolic regression with interval arithmetic and linear
  scaling.
\newblock In {\em Genetic Programming, 6th European Conference, EuroGP 2003,
  Essex, UK, April 14-16, 2003. Proceedings}, pages 70--82, 2003.

\bibitem{10.5555/138936}
J.~R. Koza.
\newblock {\em Genetic Programming: On the Programming of Computers by Means of
  Natural Selection}.
\newblock MIT Press, Cambridge, MA, USA, 1992.

\bibitem{7987733}
H.~{Li}, W.~{Luo}, and J.~{Huang}.
\newblock Localization of diffusion-based inpainting in digital images.
\newblock {\em IEEE Transactions on Information Forensics and Security},
  12(12):3050--3064, 2017.

\bibitem{Li2016ImageIA}
K.~Li, Y.~Wei, Z.~Yang, and W.~Wei.
\newblock Image inpainting algorithm based on tv model and evolutionary
  algorithm.
\newblock {\em Soft Computing}, 20:885--893, 2016.

\bibitem{6631705}
K.~{Li} and Z.~{Yang}.
\newblock Exemplar image completion based on evolutionary algorithms.
\newblock In {\em 2013 Fourth International Conference on Emerging Intelligent
  Data and Web Technologies}, pages 696--701, 2013.

\bibitem{liu2018image}
G.~Liu, F.~A. Reda, K.~J. Shih, T.~Wang, A.~Tao, and B.~Catanzaro.
\newblock Image inpainting for irregular holes using partial convolutions.
\newblock In V.~Ferrari, M.~Hebert, C.~Sminchisescu, and Y.~Weiss, editors,
  {\em Computer Vision - {ECCV} 2018 - 15th European Conference, Munich,
  Germany, September 8-14, 2018, Proceedings, Part {XI}}, volume 11215 of {\em
  Lecture Notes in Computer Science}, pages 89--105. Springer, 2018.

\bibitem{gecco2020}
L.~Manzoni, D.~Jakobovic, L.~Mariot, S.~Picek, and M.~Castelli.
\newblock Towards an evolutionary-based approach for natural language
  processing.
\newblock In C.~A.~C. Coello, editor, {\em {GECCO} '20: Genetic and
  Evolutionary Computation Conference, Canc{\'{u}}n Mexico, July 8-12, 2020},
  pages 985--993. {ACM}, 2020.

\bibitem{poli08}
R.~Poli, W.~B. Langdon, and N.~F. McPhee.
\newblock {\em A Field Guide to Genetic Programming}.
\newblock lulu.com, 2008.

\bibitem{10.1007/978-3-030-21077-9_5}
L.~Rodriguez-Coayahuitl, A.~Morales-Reyes, and H.~J. Escalante.
\newblock Convolutional genetic programming.
\newblock In J.~A. Carrasco-Ochoa, J.~F. Mart{\'i}nez-Trinidad, J.~A.
  Olvera-L{\'o}pez, and J.~Salas, editors, {\em Pattern Recognition}, pages
  47--57, Cham, 2019. Springer International Publishing.

\bibitem{ILSVRC15}
O.~Russakovsky, J.~Deng, H.~Su, J.~Krause, S.~Satheesh, S.~Ma, Z.~Huang,
  A.~Karpathy, A.~Khosla, M.~Bernstein, A.~C. Berg, and L.~Fei-Fei.
\newblock {ImageNet Large Scale Visual Recognition Challenge}.
\newblock {\em International Journal of Computer Vision (IJCV)},
  115(3):211--252, 2015.

\bibitem{schiff2011}
J.~L. Schiff.
\newblock {\em Cellular automata: a discrete view of the world}, volume~45.
\newblock John Wiley \& Sons, 2011.

\bibitem{tran2015}
C.~T. Tran, M.~Zhang, and P.~Andreae.
\newblock Multiple imputation for missing data using genetic programming.
\newblock In {\em Proceedings of the Genetic and Evolutionary Computation
  Conference, {GECCO} 2015, Madrid, Spain, July 11-15, 2015}, pages 583--590,
  2015.

\bibitem{tran2017}
C.~T. Tran, M.~Zhang, P.~Andreae, and B.~Xue.
\newblock Multiple imputation and genetic programming for classification with
  incomplete data.
\newblock In {\em Proceedings of the Genetic and Evolutionary Computation
  Conference, {GECCO} 2017, Berlin, Germany, July 15-19, 2017}, pages 521--528,
  2017.

\bibitem{10.1007/978-3-030-01264-9_1}
Z.~Yan, X.~Li, M.~Li, W.~Zuo, and S.~Shan.
\newblock Shift-net: Image inpainting via deep feature rearrangement.
\newblock In V.~Ferrari, M.~Hebert, C.~Sminchisescu, and Y.~Weiss, editors,
  {\em Computer Vision -- ECCV 2018}, pages 3--19, Cham, 2018. Springer
  International Publishing.

\end{thebibliography}

\end{document}